\documentclass[twoside]{article}

\usepackage[accepted]{aistatslike}

\usepackage[numbers,sort&compress]{natbib}

\usepackage[utf8]{inputenc}
\usepackage[T1]{fontenc}
\usepackage[hidelinks]{hyperref}
\usepackage{url}
\usepackage{booktabs}
\usepackage{amsfonts}
\usepackage{nicefrac}
\usepackage{microtype}

\usepackage{amsmath,bm,amsfonts,amssymb,amsthm,mathrsfs}
\usepackage{algorithmic}
\usepackage{bm}
\usepackage{tabularx}

  \newcommand{\eg}{\textit{e.g.}}

\newcommand{\mathbold}[1]{\bm{#1}}
\newcommand{\mbf}[1]{\mathbf{#1}}
\newcommand{\vect}[1]{\mathbf{#1}}
\newcommand{\vectb}[1]{\bm{#1}}
\newcommand{\T}{^\mathsf{T}}

\newcommand{\bigO}{\mathcal{O}}

\DeclareMathOperator{\diag}{diag}

\newcommand{\vomega}[0]{\mathbold{\omega}}

\newcommand{\MPhi}[0]{\mathbold{\Phi}}
\newcommand{\MLambda}[0]{\mathbold{\Lambda}}
\newcommand{\MSigma}[0]{\mathbold{\Sigma}}

\renewcommand{\mid}{\,|\,}

\newcommand{\vf}{\mbf{f}}

\newcommand{\vk}{\mbf{k}}

\newcommand{\vm}{\mbf{m}}

\newcommand{\vr}{\mbf{r}}

\newcommand{\vu}{\mbf{u}}

\newcommand{\vx}{\mbf{x}}
\newcommand{\vy}{\mbf{y}}
\newcommand{\vz}{\mbf{z}}

\newcommand{\MI}{\mbf{I}}

\newcommand{\MK}{\mbf{K}}

\newcommand{\MS}{\mbf{S}}

\newcommand{\MZ}{\mbf{Z}}

\usepackage{subcaption}

\usepackage{tikz,pgfplots}
\usetikzlibrary{plotmarks}

\pgfplotsset{/pgf/number format/.cd, 1000 sep={}}

\pgfplotsset{every axis/.append style={
  grid style={line width=0.6pt,dotted,gray}}}

\pgfplotsset{every axis/.append style={
  legend style={inner xsep=1pt, inner ysep=0.5pt, nodes={inner sep=1pt, text depth=0.1em},draw=none,fill=none}
}}

\pgfplotsset{every axis/.append style={
  colorbar style={width=3mm,xshift=-2mm,major tick length=2pt}
}}

\newlength{\figurewidth}
\newlength{\figureheight}

\usepackage{algorithm}
\definecolor{cgray}{gray}{0.4}

\begin{document}

\twocolumn[

\aistatstitle{Know Your Boundaries: Constraining Gaussian Processes by Variational Harmonic Features}

\aistatsauthor{ Arno Solin \And Manon Kok }

\aistatsaddress{\texttt{arno.solin@aalto.fi} \\ Department of Computer Science \\ Aalto University \And \texttt{m.kok-1@tudelft.nl} \\ Delft Center for Systems and Control \\ Delft University of Technology}]

\begin{abstract}
  Gaussian processes (GPs) provide a powerful framework for extrapolation, interpolation, and noise removal in regression and classification. This paper considers constraining GPs to arbitrarily-shaped domains with boundary conditions. We solve a Fourier-like generalised harmonic feature representation of the GP prior in the domain of interest, which both constrains the GP and attains a low-rank representation that is used for speeding up inference. The method scales as $\mathcal{O}(nm^2)$ in prediction and $\mathcal{O}(m^3)$ in hyperparameter learning for regression, where $n$ is the number of data points and $m$ the number of features. Furthermore, we make use of the variational approach to allow the method to deal with non-Gaussian likelihoods. The experiments cover both simulated and empirical data in which the boundary conditions allow for inclusion of additional physical information.
\end{abstract}

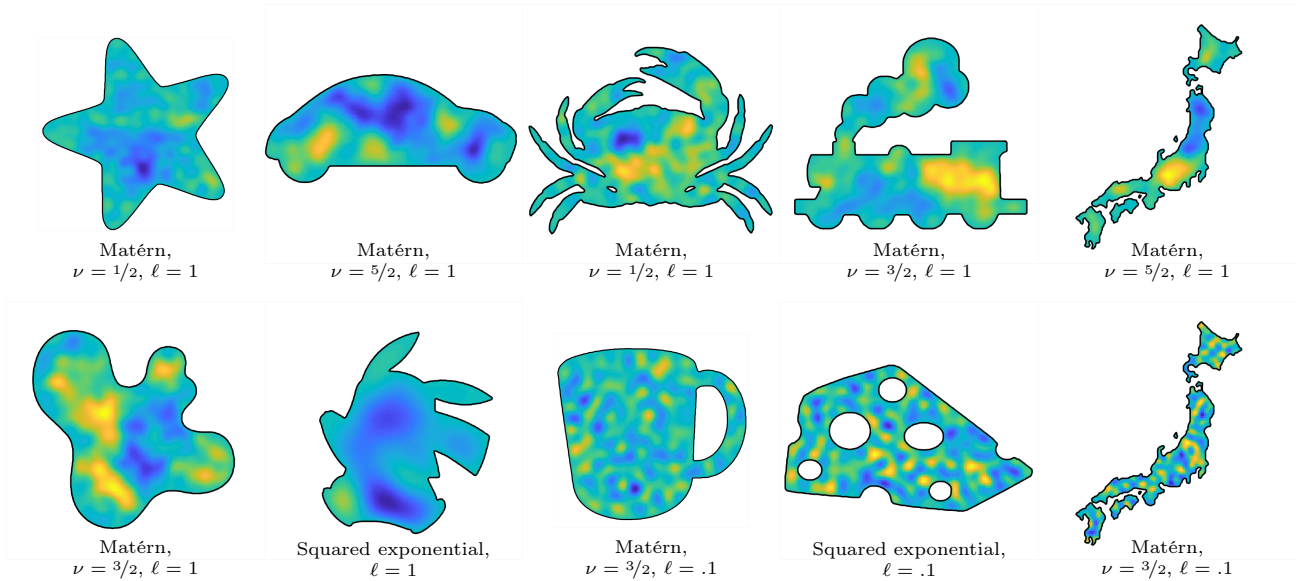
\begin{figure*}[!t]
  \centering\scriptsize

  \begin{tikzpicture}

    \node at (0\textwidth,-0.03\textwidth) 
      {\includegraphics[width=.2\textwidth]{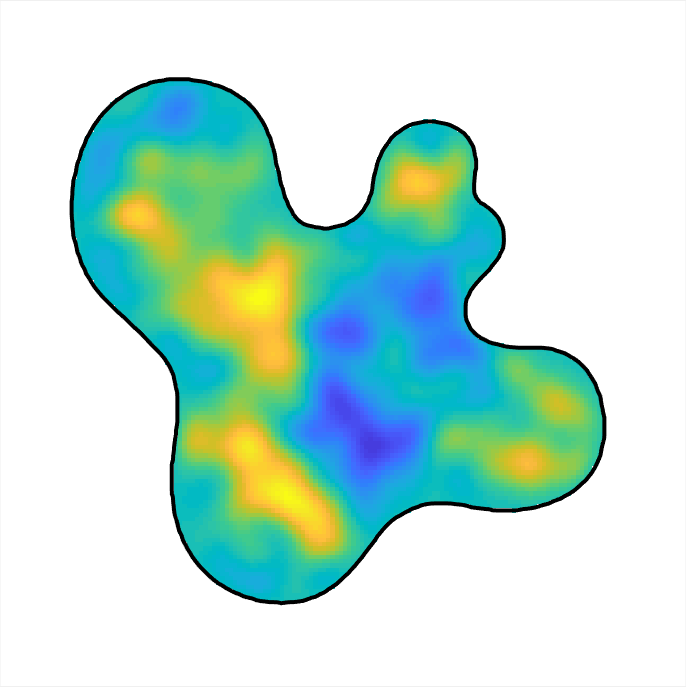}};
    \node at (0,.2\textwidth) 
      {\includegraphics[width=.15\textwidth]{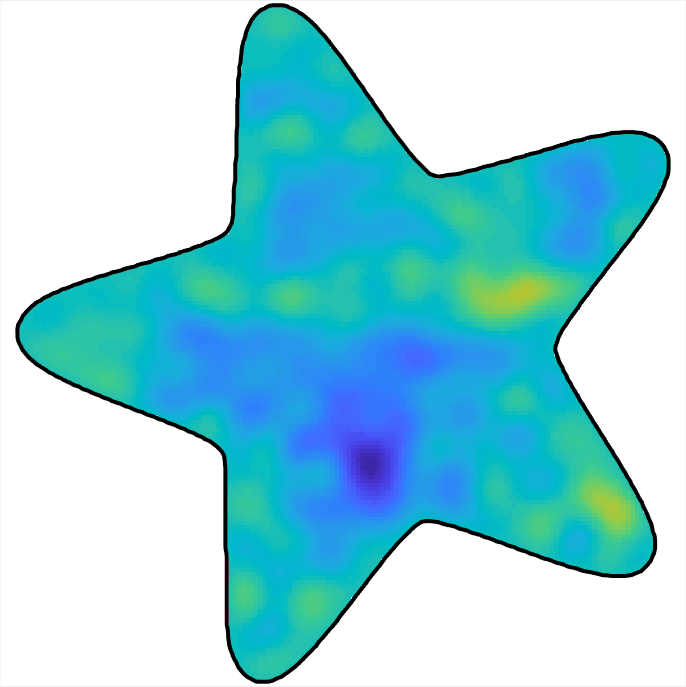}};
    \node at (.2\textwidth,.2\textwidth) 
      {\includegraphics[width=.2\textwidth]{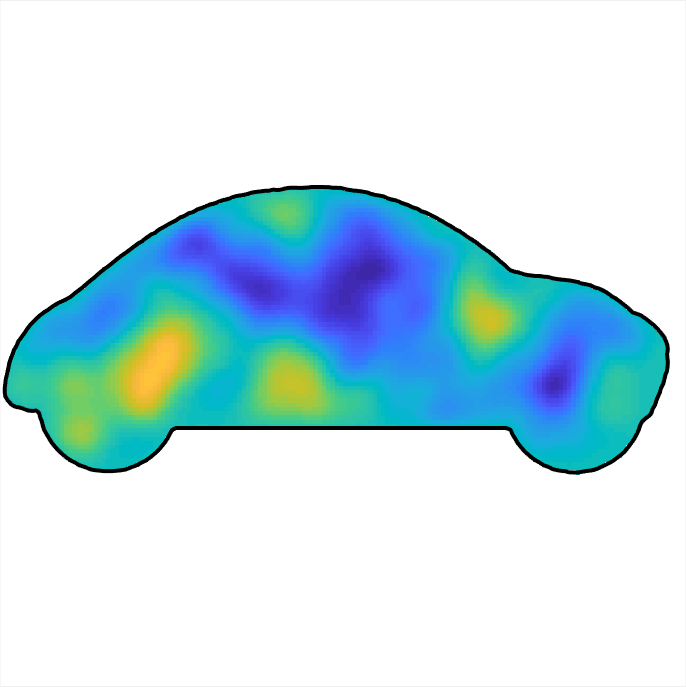}};
    \node at (.2\textwidth,-0.03\textwidth) 
      {\includegraphics[width=.2\textwidth]{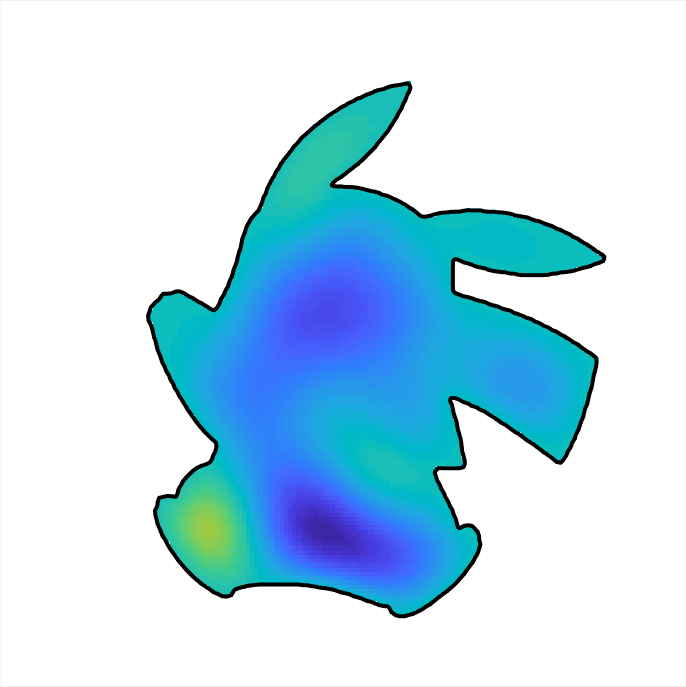}}; 
    \node at (.4\textwidth,-0.03\textwidth) 
      {\includegraphics[width=.15\textwidth]{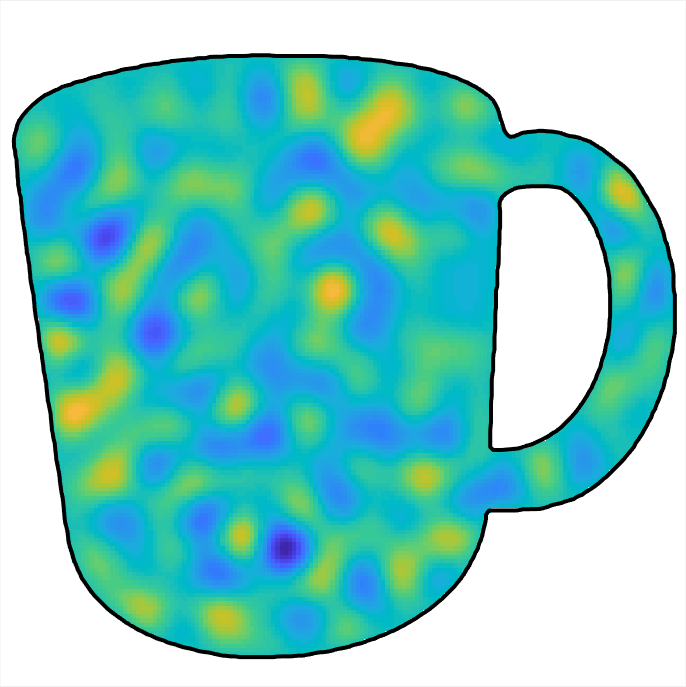}};
    \node at (.4\textwidth,.2\textwidth) 
      {\includegraphics[width=.2\textwidth]{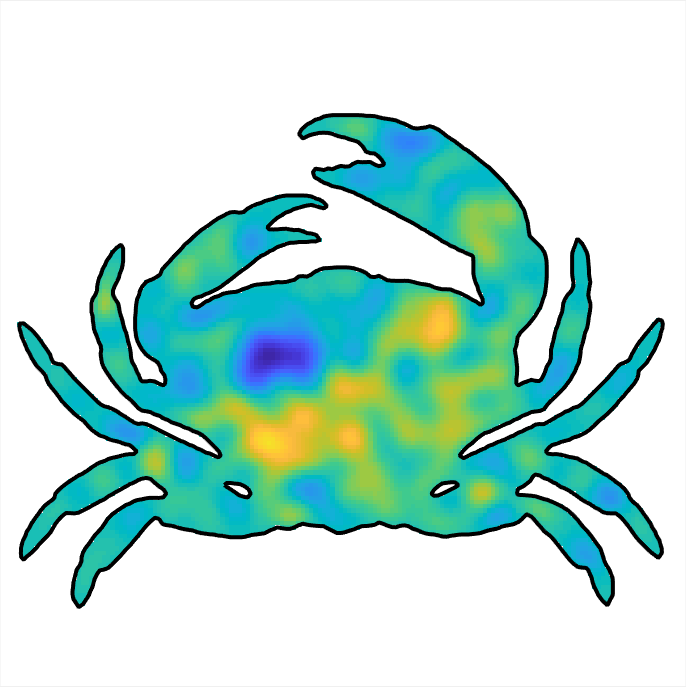}};
    \node at (.6\textwidth,.2\textwidth) 
      {\includegraphics[width=.2\textwidth]{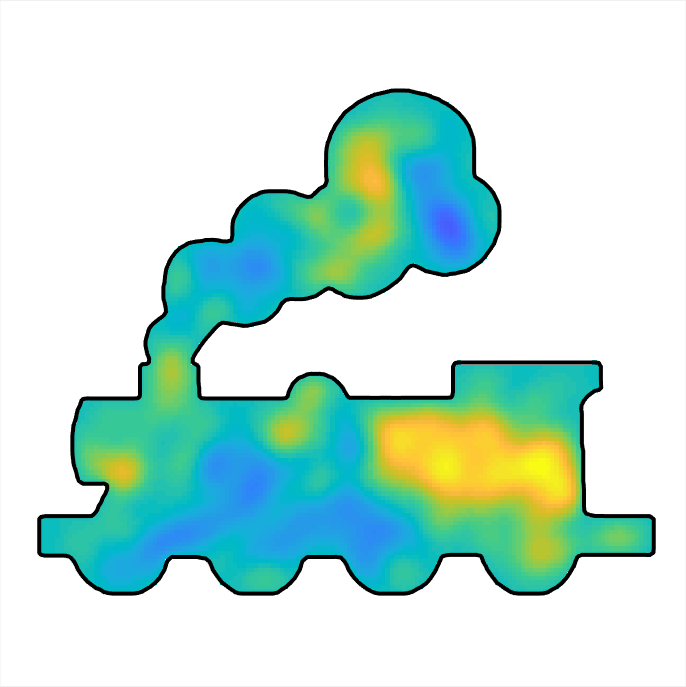}};
    \node at (.6\textwidth,-0.03\textwidth) 
      {\includegraphics[width=.2\textwidth]{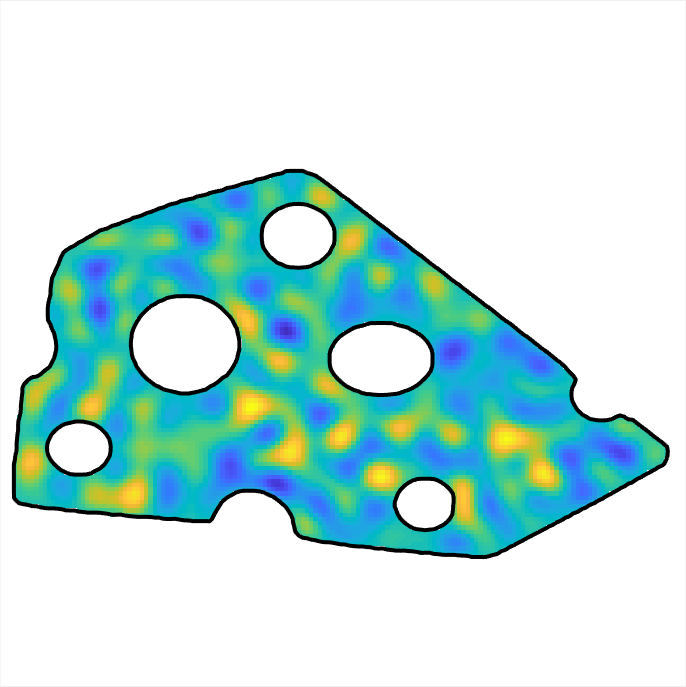}};
    \node at (.8\textwidth,.2\textwidth) 
      {\includegraphics[width=.2\textwidth]{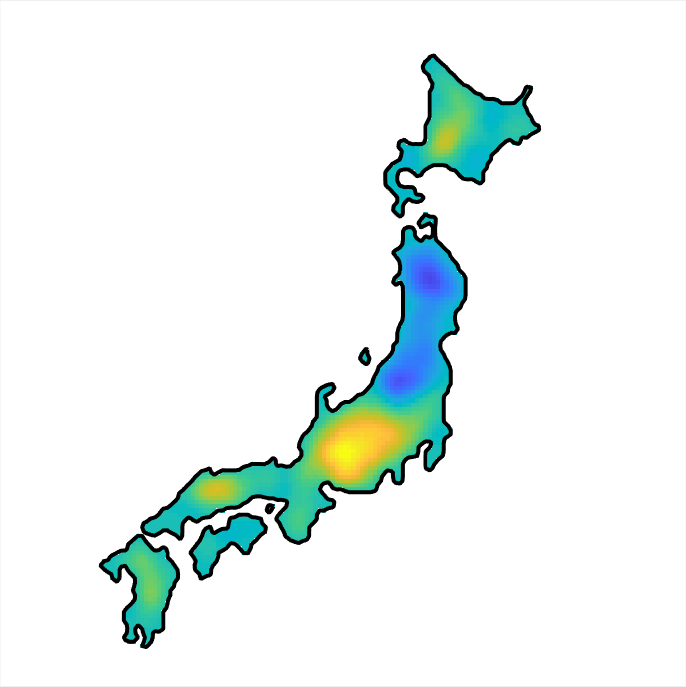}};
    \node at (.8\textwidth,-0.03\textwidth) 
      {\includegraphics[width=.2\textwidth]{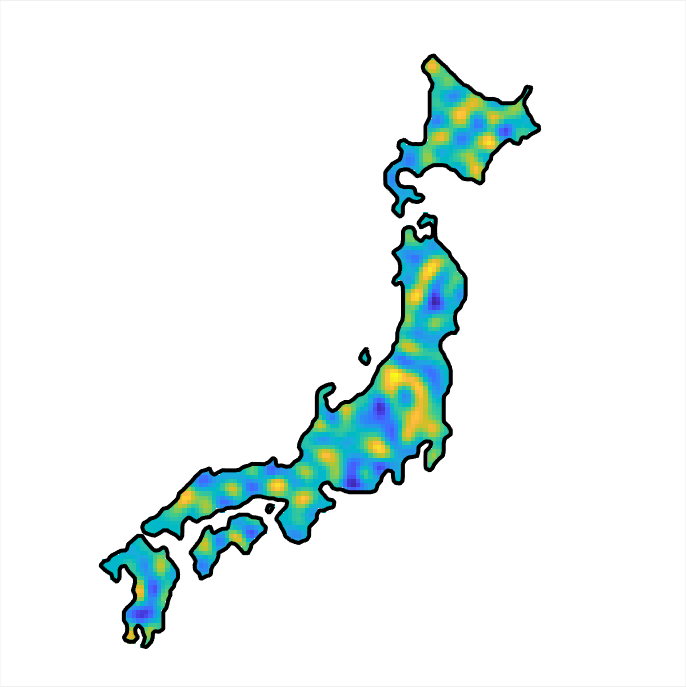}};

    \node[align=center] at (.0\textwidth,.1\textwidth) {Mat\'ern, \\ $\nu = \nicefrac{1}{2}$, $\ell=1$};
    \node[align=center] at (.0\textwidth,-.13\textwidth) {Mat\'ern, \\ $\nu = \nicefrac{3}{2}$, $\ell=1$};
    \node[align=center] at (.2\textwidth,.1\textwidth) {Mat\'ern, \\ $\nu = \nicefrac{5}{2}$, $\ell=1$};
    \node[align=center] at (.2\textwidth,-.13\textwidth) {Squared exponential, \\ $\ell=1$};
    \node[align=center] at (.4\textwidth,.1\textwidth) {Mat\'ern, \\ $\nu = \nicefrac{1}{2}$, $\ell=1$};      
    \node[align=center] at (.4\textwidth,-.13\textwidth) {Mat\'ern, \\ $\nu = \nicefrac{3}{2}$, $\ell=.1$};      
    \node[align=center] at (.6\textwidth,.1\textwidth) {Mat\'ern, \\ $\nu = \nicefrac{3}{2}$, $\ell=1$};      
    \node[align=center] at (.6\textwidth,-.13\textwidth) {Squared exponential, \\ $\ell=.1$};      
    \node[align=center] at (.8\textwidth,.1\textwidth) {Mat\'ern, \\ $\nu = \nicefrac{5}{2}$, $\ell=1$};      
    \node[align=center] at (.8\textwidth,-.13\textwidth) {Mat\'ern, \\ $\nu = \nicefrac{3}{2}$, $\ell=.1$};            
  \end{tikzpicture}
  \caption{Random draws from Gaussian process priors constrained to 2D domains of various shapes. The process goes to zero at the boundary (black line). The approach allows for non-convex and disconnected spaces. For each domain, a random draw from a GP is shown and the assigned covariance function is shown next to the domain. The scales are arbitrary and the color map is the same as in Fig.~\ref{fig:stars}.}
  \label{fig:examples}
\end{figure*}

\section{INTRODUCTION}
\label{sec:introduction}
Gaussian processes (GPs, \cite{Rasmussen+Williams:2006}) provide a widely applicable framework for probabilistic inference, where the knowledge from a measurement model (likelihood) and prior knowledge about latent functions can effectively be combined. GPs are used, for example, in robotics \cite{Ferris+Fox+Lawrence:2007, Anderson+Barfoot+Tong+Sarkka:2015, Deisenroth+Fox+Rasmussen:2015}, spatial data analysis \cite{Rue+Martino+Chopin:2009,Lindgren+Havard+Lindstrom:2011}, and signal processing \cite{PerezEtAl:2013,Kok+Solin:2018,DeisenrothEtAl:2012}.
Many applications call for including further information about the area or domain of interest, such as boundary conditions along some arbitrarily-shaped boundaries. Such constraints are often encountered in EEG/MEG brain data analysis (signals are constrained to the cortical surface), in simultaneous localisation and mapping (SLAM) in robotics (\eg, robots constrained by walls), in spatial data analysis (\eg, fish constrained to the sea), or other problems that clearly obey spatial boundaries.

Currently, the standard toolset for GPs does not include many ways for imposing  constraints from arbitrarily-shaped boundaries. One way to include this information is by introducing artificial measurements with low or zero measurement noise along the boundary. Contrary to such an artificial approach, we present a method that naturally incorporates the presence of boundary conditions for arbitrarily-shaped domains.

In previous work, box-type boundaries are typically seen as a by-product of the approximation for allowing powerful representations in terms of Fourier features~\cite{Solin+Sarkka:2014-hilbert,Hensman+Durrande+Solin:2018,Lazaro-Gredilla+Quinonero-Candela+Rasmussen+Figueiras-Vidal:2010,Fritz+Neuweiler+Nowak:2009,Paciorek:2007, John+Hensman:2018}. Typically the boundaries are chosen `far enough' from the data not to affect the results, and the main interest is in speeding up the prohibitive $\mathcal{O}(n^3)$ scaling. We also address the computational burden, but consider the presence of boundary conditions as a useful feature rather than an unfortunate necessity. One of the reasons that we can exploit the presence of boundaries is that using our method it is possible to define boundary conditions along any shape, as illustrated in Fig.~\ref{fig:examples}, rather than only along rectangular or spherical domains as in previous work. Another line of previous work are time-series motivated methods (\eg, \cite{Yang:2016}) and the schemes included in the R-INLA package \cite{Rue+Martino+Chopin:2009}; for example \cite{Lindgren+Havard+Lindstrom:2011} considers a powerful tesselation approach and addresses the inference with FEM solvers.

In GPs, apart from addressing the prohibitive computational scaling, another source of approximations are non-Gaussian likelihoods, which do not allow for closed-form inference. In this work, we use the variational approach \cite{Titsias:2009} to allow for inference in non-Gaussian likelihoods. Similarly as in \cite{Solin+Sarkka:2014-hilbert,Hensman+Durrande+Solin:2018}, we focus on problems with low-dimensional inputs.

The contributions of this paper are two-fold: 
{\em (i)}~We propose a novel approach to constrain GPs to arbitrarily-shaped domains using generalised harmonic features by extending the Hilbert space GP from~\cite{Solin+Sarkka:2014-hilbert}.
{\em (ii)}~We leverage the variational Fourier features approach from~\cite{Hensman+Durrande+Solin:2018} to allow for inference in non-Gaussian likelihoods in the constrained domains.

The paper is structured as follows. Sec.~\ref{sec:background} goes through the necessary background in GPs and variational inference. Sec.~\ref{sec:methods} delivers the methodology for the harmonic feature decomposition and the required inference formulation. In Sec.~\ref{sec:experiments} we consider both simulated data, a standard classification test data set, and a new empirical example of a constrained Log-Gaussian Cox process task. Finally, we conclude with a discussion.

\section{BACKGROUND}
\label{sec:background}
Gaussian processes (GPs, \cite{Rasmussen+Williams:2006}) are typically used as non-parametric priors over unknown functions, $f(\vx)$, and connected through a likelihood to some observations $y$. For a set of input--output pairs $\mathcal{D} = \{(\vx_i,y_i)\}_{i=1}^n$, we can write the model as
\begin{equation}\label{eq:gpModel}
  f(\vect{x})\sim\mathrm{GP}(m(\vect{x}),\kappa(\vect{x},\vect{x}')),\quad\vy\mid\vf\sim\prod_{i=1}^{n} p(y_{i}\mid f(\vect{x}_{i})).
\end{equation}
Here, the likelihood factorises over the observations. Furthermore, $m(\vect{x})$ denotes an arbitrary prior mean function and $\kappa(\vect{x},\vect{x}')$ a covariance (kernel) function. Without loss of generality we will here forth restrict the presentation to zero-mean GP priors.

A covariance function is said to be stationary if it can be written as $\kappa(\vx, \vx') \triangleq \kappa(\vx-\vx') = \kappa(\vr)$. Examples of stationary covariance functions are the squared exponential (SE) and Mat\'ern kernels given by
\begin{align}
  \kappa_{\text{SE}}(\vr) &= \sigma_\mathrm{f}^2 \exp\!\bigg( -\frac{\|\vr\|^2}{2\ell^2} \bigg),\\
  \kappa_{\text{Mat.}}(\vr) &= \sigma_\mathrm{f}^2 \frac{2^{1-\nu}}{\Gamma(\nu)}\bigg(\frac{\sqrt{2\nu}\|\vr\|}{\ell}\bigg)^\nu \mathrm{K}_\nu\bigg(\frac{\sqrt{2\nu}\|\vr\|}{\ell}\bigg), 
\end{align}
where $\sigma_\mathrm{f}^2$ and $\ell$ are magnitude and lengthscale (hyper) parameters, respectively. $\mathrm{K}_\nu(\cdot)$ is the modified Bessel function. Compared to the squared exponential kernel, the Mat\'ern kernel has an additional smoothness parameter $\nu$. Characteristics for the various degrees of smoothness, can be seen in Fig.~\ref{fig:examples}.

GP models have two drawbacks. The first is the prohibitive computational complexity which scales cubically in the number of data points. The other is the fact that the solution to a GP model is only available in closed form if the likelihood is Gaussian.
In this special case $y_i = f(\vx_i) + \varepsilon_i, \varepsilon_i \sim \mathrm{N}(0,\sigma_\mathrm{n}^2)$, the predictive mean and variance of the model functions at an unseen test input $\vx_\star$ can be written out in closed form as
\begin{align}
\begin{split} \label{eq:gp-solution-full}
  \mathbb{E}[f(\vect{x}_\star)] &= 
    \vect{k}_\star\T (\vect{K} + \sigma_\mathrm{n}^2 \vect{I}_n)^{-1} \vect{y}, \\
  \mathbb{V}[f(\vect{x}_\star)] &= 
    k(\vect{x}_\star,\vect{x}_\star) - \vect{k}_\star\T 
    (\vect{K} + \sigma_\mathrm{n}^2 \, \vect{I}_n)^{-1} \vect{k}_\star,
\end{split}
\end{align}
where $\MK_{i,j} = \kappa(\vx_i,\vx_j)$ and $\vk_{\star,i} = \kappa(\vx_\star,\vx_i)$, $i,j=1,2,\ldots,n$. Maximising the log marginal likelihood is typically employed for finding suitable hyperparameter values. That means minimising the negative log marginal likelihood w.r.t.\ $\vectb{\theta}$:
\begin{multline}\label{eq:likelihood-full}
  -\log p(\vect{y} \mid \vectb{\theta}, \mathcal{D})
  = \frac{1}{2} \log |\vect{K}_{\vectb{\theta}}+\sigma_\mathrm{n}^2\,\vect{I}_{n}| 
  \\ + \frac{1}{2} \vy\T(\vect{K}_{\vectb{\theta}} +  \sigma_\mathrm{n}^2\,\vect{I}_{n})^{-1} \vy
  + \frac{n}{2} \log(2\pi).
\end{multline}
Both Eqs.~\eqref{eq:gp-solution-full} and \eqref{eq:likelihood-full} show the problem with na\"ive GP regression as both the calculation of the determinant and the matrix inverse scale as $\mathcal{O}(n^3)$.

\subsection{Fourier Bases for GPs}
\label{sec:fourierBases}
A strategy to deal with the issue of computational complexity in GPs, is to, instead of working with the full covariance matrix, work with an approximation of it. If an eigenvalue decomposition of the covariance matrix would be available, a reduced-rank approximation $\MK \approx \MPhi \MLambda \MPhi\T$ could be made, where $\MLambda$ is a diagonal matrix of the leading $m$ eigenvalues of $\MK$ and $\MPhi$ is the matrix of the corresponding orthonormal eigenvectors. However, computing the eigenvalue decomposition is also of computational complexity $\bigO(n^3)$. One way to obtain an approximate eigenvalue decomposition in the case of stationary kernels has been presented in \cite{Solin+Sarkka:2014-hilbert}. It solves the eigendecomposition of the Laplace operator subject to Dirichlet boundary conditions on a certain domain $\Omega$ as
\begin{equation}
\begin{cases}\label{eq:eigenbasis}
-\nabla^2 \phi_j(\vect{x}) = \lambda_j^2 \phi_j(\vect{x}), 
& \vect{x} \in \Omega, \\
\phantom{-\nabla^2} \phi_j(\vect{x}) = 0, 
& \vect{x} \in \partial\Omega.
\end{cases}
\end{equation}
The eigenfunctions $\phi_j$ and eigenvalues $\lambda_j$ of the Laplace operator can be computed in closed-form for certain shapes of regular domains such as rectangles, circles, and spheres, see \cite{Solin+Sarkka:2014-hilbert}. 
The approximation of the covariance function now relies on the Fourier duality of spectral densities and covariance functions, known as the Wiener--Khintchin theorem:
\begin{alignat}{3}
    & s(\vomega) &&= \mathcal F\{k(\vr)\} &&= \int \kappa(\vr) e^{-\mathrm{i}\,\vomega\T \vr} \,\mathrm{d}\vr,\\
    & \kappa(\vr) &&= \mathcal F^{-1}\{s(\vomega)\} &&= \frac{1}{(2\pi)^d} \int s(\vomega) e^{\mathrm{i}\,\vomega\T\vr} \,\mathrm{d} \vomega \label{eq:inv_fourier}.
\end{alignat}
This theorem relies on Bochner's theorem \cite{Akhiezer+Glazman:1993}, which tells us that any continuous positive definite function, such as a covariance function, can be represented as the Fourier transform of a positive measure. If this measure has a density, it is known as the spectral density $s(\omega) \triangleq s(\|\vomega\|)$ of the covariance function. Under our convention of the Fourier transform, the spectral density corresponding to the squared exponential and Mat\'ern covariance functions are
\begin{alignat}{2}
    &s_{\text{SE}}(\omega) &&= \sigma_\mathrm{f}^2 (2 \pi \ell^2)^{\nicefrac{d}{2}} \exp\! \left( - \frac{1}{2} \omega^2 \ell^2 \right),\\
    &s_{\text{Mat.}}(\omega) &&= \sigma_\mathrm{f}^2 \frac{\Gamma(\nu + \nicefrac{d}{2})}{\Gamma(\nu)} \frac{2^d \pi^{\nicefrac{d}{2}} (2 \nu)^\nu}{\ell^{2 \nu}} \left(\frac{2 \nu}{\ell^2} + \omega^2\right)^{-\frac{2\nu+d}{2}}
\end{alignat}

where $d$ is the dimensionality of the input $\vx$. The covariance function can now be approximated as \cite{Solin+Sarkka:2014-hilbert}
\begin{equation} \label{eq:approximation}
  \kappa(\vx,\vx')
    \approx \sum_{j=1}^m s(\lambda_j) \, \phi_j(\vect{x})\,\phi_j(\vect{x}') = \MPhi \MLambda \MPhi\T, 
\end{equation}
where 
\begin{alignat}{1}
  \vectb{\Phi}_i &= 
  \begin{pmatrix}
    \phi_1(\vect{x}_i), \phi_2(\vect{x}_i), \ldots, \phi_m(\vect{x}_i) 
  \end{pmatrix}, \label{eq:eigenFun} \\
  \vectb{\Lambda} &= \diag{(s(\lambda_1), s(\lambda_2), \ldots, s(\lambda_m))}, \label{eq:eigenVal}
\end{alignat}
for $i=1,2,\ldots,n$. Similarly, we define $\vectb{\Phi}_\star$ as vectors evaluated at the prediction input location $\vect{x}_\star$. For Gaussian likelihoods, the prediction Eqs.~\eqref{eq:gp-solution-full} can now be rewritten in terms of the approximation as
\begin{align}
\begin{split} \label{eq:gp-solution-approx}
  \mathbb{E}[{f}(\vect{x}_\star)] &\approx 
    \vectb{\Phi}_\star
    (\vectb{\Phi}\T \vectb{\Phi} + \sigma_\mathrm{n}^2 \vectb{\Lambda}^{-1})^{-1}
     \vectb{\Phi}\T \vy, \\
  \mathbb{V}[{f}(\vect{x}_\star)] &\approx 
    \sigma_\mathrm{n}^2 \, \vectb{\Phi}_\star 
    (\vectb{\Phi}\T \vectb{\Phi} + \sigma_\mathrm{n}^2 \vectb{\Lambda}^{-1})^{-1}
    \vectb{\Phi}_\star\T.
\end{split}
\end{align}
For this model, the expression for evaluating the negative log marginal likelihood function for hyperparameter optimisation can be written as
\begin{align}\label{eq:likelihood-fast}
  &-\log p(\vect{y} \mid \vectb{\theta}, \mathcal{D})
  = \frac{1}{2} (n-m)\log \sigma_\mathrm{n}^2 
   + \frac{1}{2} \sum_{j=1}^m [\vectb{\Lambda}_{\vectb{\theta}}]_{j,j}  
   \nonumber \\
   &+ \frac{1}{2} \log |\sigma_\mathrm{n}^2\vectb{\Lambda}_{\vectb{\theta}}^{-1} + \vectb{\Phi}\T \vectb{\Phi}|   + \frac{n}{2} \log(2\pi)  
   \nonumber \\ 
   &+ \frac{1}{2 \sigma_\mathrm{n}^2} \big[ \vy\T\vy - \vy\T \vectb{\Phi}(\sigma_\mathrm{n}^2\vectb{\Lambda}_{\vectb{\theta}}^{-1} 
  + \vectb{\Phi}\T \vectb{\Phi})^{-1} \vectb{\Phi} \vy \big],
\end{align}

where the only remaining dependency on the covariance function hyperparameters is in the diagonal matrix $\vect{\Lambda}$ defined through the spectral density.

Since the basis functions $\MPhi$ do not depend on the hyperparameters, the product $\MPhi\T \MPhi$ can be computed once, at order $\bigO(nm^2)$, after which evaluating the likelihood is only of $\bigO(m^3)$. However, the previous work in \cite{Solin+Sarkka:2014-hilbert} does not deal with non-Gaussian likelihoods. We will therefore review some standard methods to do this in Sec.~\ref{sec:varGP}. In Sec.~\ref{sec:methods} we will then extend on the approach presented in this section and present a method that allows us to also deal with non-Gaussian likelihoods.

\subsection{Variational Gaussian Processes for General Likelihoods}
\label{sec:varGP}
Non-Gaussian likelihoods can be handled using variational approaches (see, \eg, \cite{blei2016variational}) in which the posterior is approximated by selecting the optimal distribution from a fixed family. Optimality is usually defined through the Kullback--Leibler divergence defined as (with slight abuse of notation, see also \cite{Matthews+Hensman+Turner+Ghahramani:2016})
\begin{multline}
    \textsc{KL}[q(f(\vx)) \,\|\, p(f(\vx) \mid \vy)] = \\ \mathbb E_{q(f(\vx))} \left[\log q(f(\vx)) - \log p(f(\vx) \mid \vy)\right].
    \label{eq:KL_def}
\end{multline}

The variational GP approach has extensively been used in combination with inducing inputs and has recently also been combined with Fourier-based approximations~\cite{Hensman+Durrande+Solin:2018}. In the following, we will give a short overview required for the next section, more information can for instance be found in \cite{matthews2016scalable}.
To find a family of approximating distributions $q(f(\vx))$, typically a set of inducing (pseudo) inputs $\MZ= \{ \vz_j \}_{j=1}^m$ is introduced,  where $m < n$. The values of the function at $\MZ$ are denoted by $\vu = \{ \vu_j \}_{j = 1}^m$. Since $f(\vx)$ and $\vu$ are jointly Gaussian, it is possible to write the function $f(\vx)$ conditioned on the values $\vu$ as
\begin{multline}
    f(\vx) \mid \vu \sim \mathrm{GP}\left(\vk_\mathrm{u}(\vx)\T \MK_{\mathrm{uu}}^{-1}\vu, \right. \\
     \left. \kappa(\vx, \vx') - \vk_\mathrm{u}(\vx)\T \MK_{\mathrm{uu}}^{-1}\vk_\mathrm{u}(\vx')\right).
\label{eq:gp_cond_sparse}
\end{multline}
The joint approximation for the posterior is $q(\vu)\,q(f(\vx)\mid \vu)$. We therefore must also choose some approximate posterior distribution $q(\vu)$, whose exact form will depend on the likelihood $p(\vy \mid f(\vx))$. This can be done by minimising the Kullback--Leibler divergence~\eqref{eq:KL_def}. Expanding the true posterior using Bayes' rule, \eqref{eq:KL_def} can be written as
\begin{align}
  &\textsc{KL}\left[q(f(\vx))\,\|\,p(f(\vx)\mid\vy)\right] \nonumber \\
  &= -\mathbb E_{q(f(\vx))}\left[ \log \frac{p(\vy\mid f(\vx)) \, p(f(\vx))}{q(f(\vx))}\right] + \log p(\vy)\nonumber \\
  &\triangleq -\textsc{ELBO} + \log p(\vy)\,.
    \label{eq:elbo}
\end{align}
Minimising the Kullback--Leibler objective is therefore equivalent to maximising the Evidence Lower Bound (ELBO).

We factor $p(f(\vx)) = p(\vu)\,p(\vf\mid \vu)\,p(f \mid \vf,\vu)$ with
\begin{align}
    p(\vu) &= \mathrm N\left(\mathbf{0},\, \MK_\mathrm{uu}\right), \nonumber \\
    p(\vf\mid\vu) &= \mathrm N\left(\MK_\mathrm{fu}\MK_\mathrm{uu}^{-1}\vu,\, \MK_\mathrm{ff} - \MK_\mathrm{fu}\MK_\mathrm{uu}^{-1}\MK_\mathrm{fu}\T\right), \nonumber \\
    p(f(\vx)\mid \vf,\vu) &= \mathrm{GP}\left(m^\star(\vx),\,\kappa^\star(\vx, \vx')\right),
\end{align}
where $m^\star(\vx)$ and $k^\star(\vx, \vx')$ are the usual Gaussian process conditional mean and variance, conditioned on both $\vf$ and $\vu$.
The approximate posterior process can be factored similarly as
\begin{align}
    q(\vf\mid\vu) &= \mathrm N\left(\MK_\mathrm{fu}\MK_\mathrm{uu}^{-1}\vu,\, \MK_\mathrm{ff} - \MK_\mathrm{fu}\MK_\mathrm{uu}^{-1}\MK_\mathrm{fu}\T\right), \nonumber \\
    q(f(\vx)\mid \vf,\vu) &= \mathrm{GP}\left(m^\star(\vx),\,\kappa^\star(\vx, \vx')\right). 
\end{align}
Since the two processes $p(f(\vx))$ and $q(f(\vx))$ are the same except for $p(\vu)$ and $q(\vu)$, the ELBO \eqref{eq:elbo} simplifies to
\begin{multline}
    \textsc{ELBO} = \mathbb E_{q(\vu)q(\vf\mid\vu)} \big[\log p(\vy\mid\vf)\big] \\ -
    \mathbb E_{q(\vu)}\left[\log\frac{q(\vu)}{p(\vu)}\right].
    \label{eq:gp_elbo}
\end{multline}

It is possible to show that the distribution $\hat q(\vu)$ that maximises the ELBO is given by
\begin{equation}
    \log \hat q(\vu) = \mathbb E_{q(\vf\mid \vu)} \left[\log p(\vy\mid\vf)\right] + \log p(\vu) + \text{const.}
    \label{eq:q_hat}
\end{equation}
This distribution is intractable for general likelihoods, but can be approximated using a Gaussian distribution~\cite{Hensman+Matthews+Ghahramani:2015}. Maximising the ELBO then corresponds to optimising over the mean and the covariance of this Gaussian. If the approximating Gaussian distribution $q(\vu)$ has mean $\vm$ and covariance $\MSigma$, then the entire approximating process is a GP, with
\begin{multline}
    q(f(\vx)) = \int q(\vu)\,q(f(\vx)\mid \vu)\,\mathrm{d} \vu  \\
            = \mathrm {GP}\left(\vk_\mathrm{u}(\vx)\T \MK_\mathrm{uu}^{-1}\vm, \, \kappa(\vx, \vx') +  \right. \\
             \left. \vk_\mathrm{u}(\vx)\T (\MK_\mathrm{uu}^{-1}\MSigma\MK_\mathrm{uu}^{-1} - \MK_\mathrm{uu}^{-1})\vk_\mathrm{u}(\vx')\right).
\label{eq:gp_pred_sparse}
\end{multline}
For the special case where the data-likelihood $p(y_i\mid f(\vx_i))$ is Gaussian with noise-variance $\sigma_\mathrm{n}^2$, the optimal distribution for $q(\vu)$ is given by~\cite{Hensman+Fusi+Lawrence:2013}
\begin{equation}
\begin{aligned}
    \hat{q}(\vu) &= \mathrm N\left(\hat{\vm},\, \hat{\MSigma} \right), \\
  \hat{\MSigma} &= [\MK_\mathrm{uu}^{-1} + \sigma_\mathrm{n}^{-2}\MK_\mathrm{uu}^{-1}\MK_\mathrm{uf}\MK_\mathrm{uf}\T\MK_\mathrm{uu}^{-1}]^{-1}, \\
  \hat{\vm}&=\sigma_\mathrm{n}^{-2} \hat{\MSigma} \MK_\mathrm{uu}^{-1} \MK_\mathrm{uf}\vy,
\end{aligned}
\label{eq:optimal_q}
\end{equation}
and the \textsc{ELBO} at this optimal point is
\begin{multline}
    \textsc{ELBO}\left(q\right) = \log \mathrm N\left(\vy \mid 0,\, \MK_\mathrm{fu}\MK_\mathrm{uu}^{-1}\MK_\mathrm{uf} + \sigma_\mathrm{n}^2 \MI\right) \\ - 
    \frac{1}{2}\sigma_\mathrm{n}^{-2} \mathrm{tr}\!\left(\MK_\mathrm{ff} - \MK_\mathrm{fu}\MK_\mathrm{uu}^{-1}\MK_\mathrm{uf}\right).
\end{multline}

\section{METHODS}
\label{sec:methods}
In this paper, the main interest is in including boundary conditions in the standard GP model~\eqref{eq:gpModel} resulting in 
\begin{equation}
\begin{aligned}
  f(\vect{x}) &\sim\mathrm{GP}(0,\kappa(\vx,\vx')), \\
  \text{s.t.} ~ & f(\vx) = 0, \quad \vx \in \partial \Omega, \\
  \vy \mid \vf &\sim\prod_{i=1}^{n} p(y_{i} \mid f(\vx_{i})),
\end{aligned}
\label{eq:gpModel-bound}
\end{equation}
where $\Omega \subset \mathbb{R}^d$ is the domain of interest and $\partial\Omega$ denotes its boundary. This means that {\it a~priori} we assume the GP to be generated by a GP prior with stationary covariance function $\kappa(\vx,\vx')$ with the additional constraints of smoothly attaining the given boundary constraint. Thus the effective covariance function in the domain-aware GP prior is highly non-stationary.

Note that without loss of generality we can change the boundary constraint to be constant. Other boundary conditions, such as Neumann conditions (derivative going to zero at boundary) can also be included. Since we no longer assume a rectangular or spherical domain, it is no longer possible, like in Sec.~\ref{sec:fourierBases}, to compute the eigendecomposition of the Laplace operator in closed form. In this section we will now first discuss how to compute the harmonic features numerically. We will then discuss how the variational approach from Sec.~\ref{sec:varGP} can be used to handle non-Gaussian likelihoods. 

\begin{figure}[!t]
  \centering\scriptsize
  \setlength{\figurewidth}{0.1\textwidth}
  \setlength{\figureheight}{\figurewidth}
  \begin{subfigure}[b]{.49\columnwidth}
    \includegraphics[width=\columnwidth]{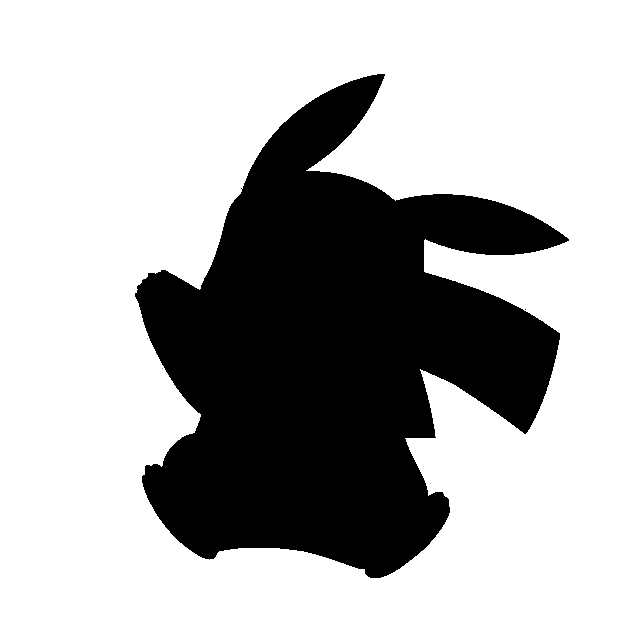}
    \caption{Domain}
    \label{fig:domain}
  \end{subfigure}
  \begin{subfigure}[b]{.49\columnwidth}
    \pgfplotsset{yticklabel style={rotate=90}, ylabel style={yshift=-15pt},clip=true,scale only axis,axis on top,clip marker paths,legend style={row sep=0pt},legend columns=-1,xlabel near ticks,legend style={fill=white}}
    \setlength{\figurewidth}{.8\textwidth}
    \setlength{\figureheight}{\figurewidth}
%
%
\begin{tikzpicture}

\begin{axis}[%
axis on top,
xmin=-5.83941605839416,
xmax=8005.83941605839,
y dir=reverse,
ymin=-5.46418128654971,
ymax=7480.46418128655,
axis background/.style={fill=white},
legend style={legend cell align=left,align=left,draw=white!15!black},
width=\figurewidth,
height=\figureheight
]
\addplot [forget plot] graphics [xmin=-5.83941605839416,xmax=8005.83941605839,ymin=-5.46418128654971,ymax=7480.46418128655] {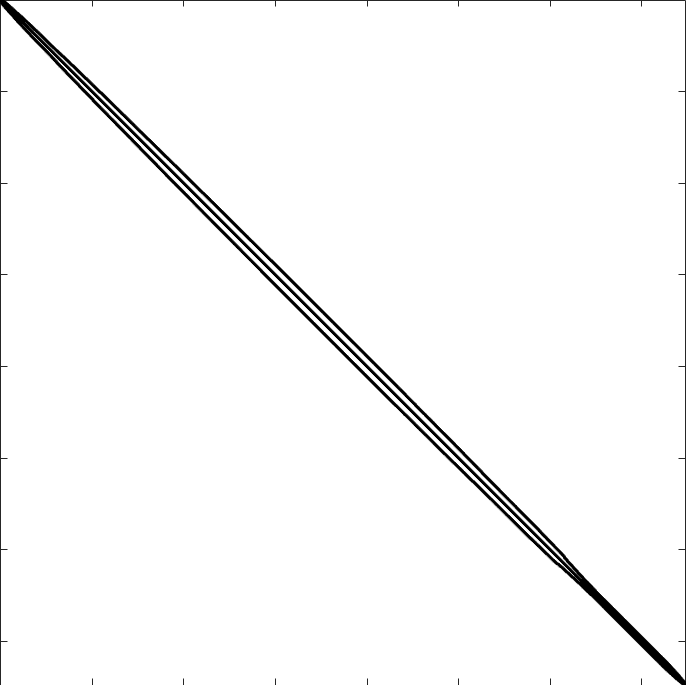};
\end{axis}
\end{tikzpicture}%
    \caption{Sparse stencil matrix}
    \label{fig:stencil}
  \end{subfigure}
  \\[1em]
  \begin{subfigure}{\columnwidth}  
  \begin{tikzpicture}
    \foreach \i in {1,...,15} {
      \node[draw=white,fill=black!20,minimum size=\figurewidth,inner sep=0pt]
        (\i) at ({\figurewidth*mod(\i-1,5)},{-\figureheight*int((\i-1)/5)})
        {\includegraphics[width=\figurewidth]{./fig/basis/basis-\i.png}};
      \node at ({\figurewidth*mod(\i-1,5)-.3\figurewidth},
                {-\figureheight*int((\i-1)/5)+.3\figureheight})
        {\i};
      };
  \end{tikzpicture}
  \caption{Harmonic basis functions}
  \label{fig:basis}
  \end{subfigure}
  \caption{Example domain for which we can numerically compute the harmonic basis functions using the sparse stencil matrix.}
  \label{fig:domain-stencil-basis}
\end{figure}

\subsection{Computing the Harmonic Features}
\label{sec:numEig}
Instead of computing the eigendecomposition of the Laplace operator in closed form as in Sec.~\ref{sec:fourierBases}, we solve the eigendecomposition numerically. We first turn our domain of interest into a grid mask (\eg, in Fig.~\ref{fig:domain-stencil-basis} we use a $162{\times}162$ grid). We then approximate the Laplacian using a 9-point finite difference approximation with step size $h$ (determined by the physical size of the  domain) as 
\begin{align}
&-\nabla^2 u(x_1,x_2) \nonumber \\
& {\scriptstyle\approx \frac{1}{h^2} \big[ 
  \frac{2}{3} u(x_1{+}h,x_2) {+} \frac{2}{3} u(x_1{-}h,x_2) {+} \frac{2}{3} u(x_1,x_2{+}h) }\nonumber \\ 
  &{\scriptstyle+ \frac{2}{3} u(x_1,x_2{-}h) {+} \frac{1}{3} u(x_1{+}h,x_2{+}h) {+} \frac{1}{3} u(x_1{+}h,x_2{-}h)} \nonumber \\ 
  &{\scriptstyle+ \frac{1}{3} u(x_1{-}h,x_2{+}h) {+} \frac{1}{3} u(x_1{-}h,x_2{-}h) - \frac{10}{3} u(x_1,x_2) \big]}.
\label{eq:9pointDiff}
\end{align}
This operation can be written as an operation by a sparse {\em stencil} matrix $\MS$. By letting this stencil matrix only work on locations that are inside the domain $\Omega$, the boundary conditions can naturally be included. Let us consider an irregularly-shaped domain $\Omega$, for instance the one displayed in Fig.~\ref{fig:domain}. This results in a stencil matrix (of size $162^2{\times}162^2$) with the sparsity pattern displayed in Fig.~\ref{fig:stencil}. The number of non-zero entries in this stencil matrix is 65,596 which corresponds to $\sim$1\textperthousand~of the elements.

We form the stencil corresponding to the domain of interest and solve the $m$ largest, real eigenvalues $\lambda_{j}^2$ and the corresponding eigenvectors $\phi_{j}(\vect{x})$ of the stencil matrix using a Krylov--Schur algorithm~\cite{Stewart:2002,Lehoucq+Sorensen+Yang:1998}. The implementation is part of ARPACK (\url{https://www.caam.rice.edu/software/ARPACK/}) and callable in \textsc{Matlab} as \texttt{eigs} and in Python  through {\em scipy.sparse.linalg.eigs}. The first $25$ harmonic basis functions of the example domain from Fig.~\ref{fig:domain} are shown in Fig.~\ref{fig:basis}.

Note that using a Taylor expansion of the different terms in~\eqref{eq:9pointDiff}, it can be shown that in~\eqref{eq:9pointDiff} we actually compute $-\nabla^2 u(x_1,x_2) - \frac{h^2}{12} \nabla^4 u(x_1,x_2) + \mathcal{O}(h^4)$. Ignoring these higher order terms, the eigenvalue problem that is being solved is therefore instead
\begin{align}
- \nabla^2 u(x_1,x_2) - \frac{h^2}{12} \nabla^4 u(x_1,x_2) = \lambda^2 u(x_1,x_2).
\end{align}
The eigenvalues can be corrected for this as 
\begin{align}
  \bar{\lambda}_{j}^2 = \left.{2 \lambda_j^2}\middle/{\sqrt{1 + \tfrac{\lambda_j^2 h^2}{3}} + 1}\right. .
\end{align}

\begin{figure*}[!t]
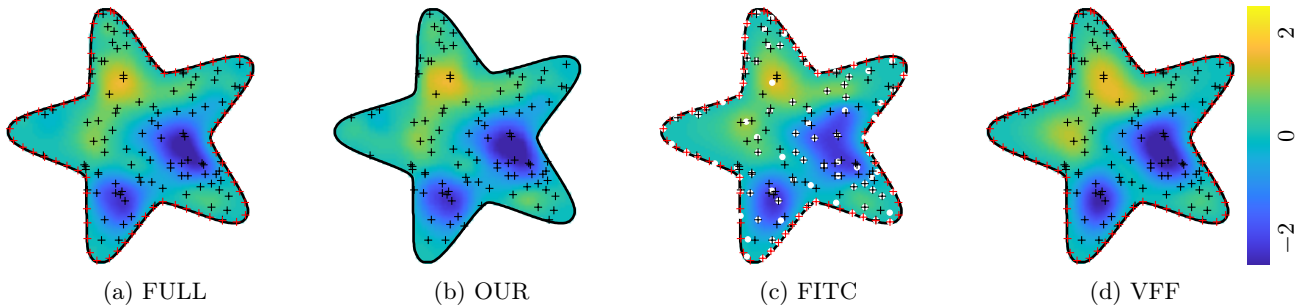


  \footnotesize
  \pgfplotsset{yticklabel style={rotate=90}, ylabel style={yshift=-15pt},clip=true,scale only axis,axis on top,clip marker paths,legend style={row sep=0pt},legend columns=-1,xlabel near ticks,legend style={fill=white},axis line style={draw=none},tick style={draw=none}}
  \setlength{\figurewidth}{.20\textwidth}
  \setlength{\figureheight}{\figurewidth}
  \begin{subfigure}[b]{.24\textwidth}
    \input{./fig/regression-full.tex}
    \caption{FULL}
  \end{subfigure}
  \hspace*{\fill}
  \begin{subfigure}[b]{.24\textwidth}
    \input{./fig/regression-our.tex}
    \caption{OUR}
  \end{subfigure}
  \hspace*{\fill}
  \begin{subfigure}[b]{.24\textwidth}
    \input{./fig/regression-fitc.tex}
    \caption{FITC}
  \end{subfigure}
  \hspace*{\fill}
  \begin{subfigure}[b]{.24\textwidth}
    \input{./fig/regression-vff.tex}
    \caption{VFF}
  \end{subfigure}
  \caption{Example results for the regression example in a star-shaped domain. The boundary (black line) enforces the process to go to zero. The black crosses are the training inputs. In the comparison methods, the red crosses are the noise-free boundary measurements. In (c), the white dots are the inducing inputs. Even in this simple example, (c) and (d) have difficulties predicting the left-side arm.}
  \label{fig:stars}
\end{figure*}

\subsection{Variational Harmonic Features}
\label{sec:VHF}
The variational approach from Sec.~\ref{sec:varGP} can be used for variational inference in a model in which the Fourier features are either computed in closed-form as in Sec.~\ref{sec:fourierBases} or numerically as in Sec.~\ref{sec:numEig}. The harmonic features will hereby play a similar role as the inducing points in Sec.~\ref{sec:varGP}. Assume that our function values are defined in terms of our features as $f(\vx) = \MPhi(\vx) \vu$ and that the prior over these features is $p(\vu) = \mathrm{N}(\vect{0},\MLambda)$, where $\MPhi(\vx)$ and $\MLambda$ are defined in~\eqref{eq:eigenFun} and \eqref{eq:eigenVal}, respectively. Similar to \eqref{eq:gp_cond_sparse}, the function values and inducing inputs are then jointly Gaussian. We will now follow a similar approach to the one taken in Sec.~\ref{sec:varGP} and approximate the posterior by minimising the Kullback--Leibler divergence. Hence, we are again interested in computing the posterior $q(\vu)$ that maximises the ELBO defined in~\eqref{eq:gp_elbo}.

For the special case of a Gaussian likelihood, the posterior $q(\vu)$ is again given by~\eqref{eq:optimal_q}. Using the relation that for our model $\MK_\mathrm{uu} = \MLambda$ and $\MK_\mathrm{fu} = \MPhi \MLambda$, we can write the posterior $q(\vu)$ as
\begin{align}
    \hat q(\vu) = \mathrm N\big(\hat\vm,\, \hat\MSigma \big), & &
  \hat\MSigma &= [\MLambda^{-1} + \sigma_n^{-2} \MPhi\T \MPhi]^{-1}, \nonumber \\
  & &\hat\vm &= \sigma_n^{-2}\hat\MSigma \MLambda \MPhi\T \vy.
\label{eq:optimal_q_harm}
\end{align}
Using the relation $f(\vx) = \MPhi(\vx) \vu$, the posterior over the function values $\vf$ is therefore given by 
\begin{align}
    \vf \sim \mathrm N\big(\hat\vm_\mathrm{f},\, \hat\MSigma_\mathrm{f} \big), 
  & & \hat\MSigma_\mathrm{f} &= \MPhi [\MLambda^{-1} + \sigma_n^{-2} \MPhi\T \MPhi]^{-1} \MPhi\T, \nonumber \\
  & & \hat\vm_\mathrm{f} &= \sigma_n^{-2} \MPhi \hat\MSigma_\mathrm{f} \MPhi\T \vy.
\label{eq:optimal_q_harm_f}
\end{align}
Note that these equations exactly correspond to~\eqref{eq:gp-solution-approx} presented in Sec.~\ref{sec:fourierBases} if these are used to predict on $\vx$. Using this relation between the methods, it is now possible to use the variational inference scheme from Sec.~\ref{sec:varGP} together with the harmonic features derived in Sec.~\ref{sec:numEig} to do inference in the model~\eqref{eq:gpModel-bound} for arbitrary boundary conditions and general likelihoods.

\section{EXPERIMENTS}
\label{sec:experiments}
We include three different experiments showing the properties of our method. The first example is a simulation study that compares our method to alternative solutions in a regression setup. The second example considers the {\em banana} classification dataset with a hard decision boundary. The third example is a new empirical example of modelling tick bite density.

The methods in Sec.~\ref{sec:methods} were implemented in both Mathworks \textsc{Matlab} and Python. The codes for replicating them are available at \url{https://github.com/AaltoML/boundary-gp}. We leveraged the GP machinery available in TensorFlow~\cite{tensorflow} and GPflow~\cite{matthews2017gpflow}. GPflow was also used for the comparison methods.

\subsection{Benchmarking}
\label{sec:benchmarking}
We consider a simulated setup where we choose a star-shaped domain of unit width and a Dirichlet boundary condition enforcing the process to be zero at the boundary. We simulate 10 data sets which obey the assumption of the process going to zero at the boundary,  observe $n=100$ uniformly random input locations in the domain (black crosses in Fig.~\ref{fig:stars}), and add Gaussian noise to the measurements.

As a baseline, we solve the GP regression problem na\"ively by including a number of noise-free observations along the boundary. We use 73 points along the boundary (red crosses in Fig.~\ref{fig:stars}) which are considered as normal measurements in the GP regression problem, but with a noise scale of zero.
We compare our approach to two general-purpose schemes for rank-reduced GP regression: The (gold-standard) {\em Fully independent training conditional} (FITC) method (see~\cite{Quinonero-Candela+Rasmussen:2005}) and {\em Variational Fourier features} (VFF, \cite{Hensman+Durrande+Solin:2018}). With the expectation that increasing the number of inducing points should improve every method, we applied 4 to 100 inducing points/features. For all models, we used a Gaussian process prior with a Mat\'ern ($\nu=\nicefrac{3}{2}$) covariance function. In the FULL, FITC, and our method the standard non-separable covariance function was used, while VFF decomposed the covariance function to a product of two Mat\'erns over the different input dimensions. This Kronecker structure in VFF needs less inducing features, and including too many features leads to instability in evaluation of the model, as previously discussed in \cite{Hensman+Durrande+Solin:2018} (thus $m=100$ is not included for VFF).
For FITC and VFF, the noise-free inputs were problematic for hyperparameter and inducing input location optimisation, resulting in numerical instability. These problems were not encountered in the FULL and proposed model. For a fair comparison of representative power, we hence fixed the hyperparameters ($\sigma_\mathrm{f}^2 = 1$, $\ell = 0.1$, $\sigma_\mathrm{n}^2 = 0.1^2$) for all models and chose the inducing input locations for FITC by $k$-means clustering.

Fig.~\ref{fig:stars} shows the predictive mean for one of the data sets with the four different methods ($m=64$). The na\"ive FULL GP with the noise-free observations clearly best agrees with the proposed method that can directly use the boundary information as part of the GP prior. FITC and VFF do a fair job but differ even visually from the results in (a)--(b). Clear differences can be seen in the blue middle part and in the left arm.

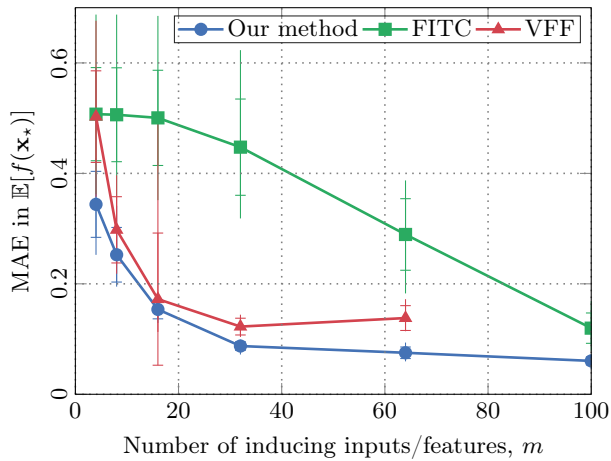
\begin{figure}[!t]
  \centering\footnotesize
  \pgfplotsset{yticklabel style={rotate=90}, ylabel style={yshift=-15pt},clip=true,scale only axis,axis on top,clip marker paths,legend style={row sep=0pt},legend columns=-1,xlabel near ticks,legend style={fill=white}}
  \setlength{\figurewidth}{.4\textwidth}
  \setlength{\figureheight}{.75\figurewidth}
%
%
\definecolor{mycolor1}{rgb}{0.26667,0.44706,0.70980}%
\definecolor{mycolor2}{rgb}{0.16471,0.67059,0.38039}%
\definecolor{mycolor3}{rgb}{0.82745,0.26275,0.30588}%
\begin{tikzpicture}

\begin{axis}[%
unbounded coords=jump,
xmin=0,
xmax=100,
xlabel={Number of inducing inputs/features, $m$},
xmajorgrids,
ymin=0,
ymax=0.7,
ylabel={MAE in $\mathbb{E}[f(\vx_\star)]$},
ymajorgrids,
axis background/.style={fill=white},
legend style={legend cell align=left,align=left,draw=white!15!black},
width=\figurewidth,
height=\figureheight
]
\addplot [color=mycolor1,solid,forget plot]
 plot [error bars/.cd, y dir = both, y explicit]
 table[row sep=crcr, y error plus index=2, y error minus index=3]{%
4	0.343901637678329	0.0597454060067256	0.0597454060067256\\
8	0.252648968468275	0.0494318042035517	0.0494318042035517\\
16	0.153671466464563	0.0170664493168746	0.0170664493168746\\
32	0.087364650404349	0.0080575904094229	0.0080575904094229\\
64	0.0751441681669349	0.0104699474697263	0.0104699474697263\\
100	0.0604301023475071	0.00751002236156257	0.00751002236156257\\
};
\addplot [color=mycolor1,solid,forget plot]
  table[row sep=crcr]{%
4	0.427241610507218\\
4	0.252479313316524\\
nan	nan\\
};
\addplot [color=mycolor1,solid,forget plot]
  table[row sep=crcr]{%
8	0.341242986480901\\
8	0.194918626057857\\
nan	nan\\
};
\addplot [color=mycolor1,solid,forget plot]
  table[row sep=crcr]{%
16	0.180901226632028\\
16	0.129171727539953\\
nan	nan\\
};
\addplot [color=mycolor1,solid,forget plot]
  table[row sep=crcr]{%
32	0.0990568340234287\\
32	0.0720060343254972\\
nan	nan\\
};
\addplot [color=mycolor1,solid,forget plot]
  table[row sep=crcr]{%
64	0.0936237465107281\\
64	0.0596241064956288\\
nan	nan\\
};
\addplot [color=mycolor1,solid,forget plot]
  table[row sep=crcr]{%
100	0.0717793198357129\\
100	0.0451527897302432\\
nan	nan\\
};
\addplot [color=mycolor2,solid,forget plot]
 plot [error bars/.cd, y dir = both, y explicit]
 table[row sep=crcr, y error plus index=2, y error minus index=3]{%
4	0.50721799226367	0.0843381876610691	0.0843381876610691\\
8	0.506164094763802	0.0850162667719082	0.0850162667719082\\
16	0.500499716706042	0.0862006515841376	0.0862006515841376\\
32	0.447455552832308	0.0871479639804109	0.0871479639804109\\
64	0.289368086247984	0.0647975741635679	0.0647975741635679\\
100	0.119744125730974	0.0274689371368849	0.0274689371368849\\
};
\addplot [color=mycolor2,solid,forget plot]
  table[row sep=crcr]{%
4	0.687530397136995\\
4	0.362169292233756\\
nan	nan\\
};
\addplot [color=mycolor2,solid,forget plot]
  table[row sep=crcr]{%
8	0.687470080320008\\
8	0.358493327689662\\
nan	nan\\
};
\addplot [color=mycolor2,solid,forget plot]
  table[row sep=crcr]{%
16	0.685135869388468\\
16	0.351536847776481\\
nan	nan\\
};
\addplot [color=mycolor2,solid,forget plot]
  table[row sep=crcr]{%
32	0.623137225271308\\
32	0.318414758895302\\
nan	nan\\
};
\addplot [color=mycolor2,solid,forget plot]
  table[row sep=crcr]{%
64	0.387207872160631\\
64	0.182658681369279\\
nan	nan\\
};
\addplot [color=mycolor2,solid,forget plot]
  table[row sep=crcr]{%
100	0.168579394313346\\
100	0.0788934863517566\\
nan	nan\\
};
\addplot [color=mycolor3,solid,forget plot]
 plot [error bars/.cd, y dir = both, y explicit]
 table[row sep=crcr, y error plus index=2, y error minus index=3]{%
4	0.502773073223379	0.082927666808751	0.082927666808751\\
8	0.29778784064398	0.0599821730366659	0.0599821730366659\\
16	0.172286157081449	0.119654998247515	0.119654998247515\\
32	0.122560631964565	0.0151700658138649	0.0151700658138649\\
64	0.138024997424892	0.0224852088517508	0.0224852088517508\\
100	nan	nan	nan\\
};
\addplot [color=mycolor3,solid,forget plot]
  table[row sep=crcr]{%
4	0.676413040038175\\
4	0.357594486828519\\
nan	nan\\
};
\addplot [color=mycolor3,solid,forget plot]
  table[row sep=crcr]{%
8	0.396567764269392\\
8	0.21822398659813\\
nan	nan\\
};
\addplot [color=mycolor3,solid,forget plot]
  table[row sep=crcr]{%
16	0.511090114992461\\
16	0.11288419904574\\
nan	nan\\
};
\addplot [color=mycolor3,solid,forget plot]
  table[row sep=crcr]{%
32	0.144110129018186\\
32	0.103779834355489\\
nan	nan\\
};
\addplot [color=mycolor3,solid,forget plot]
  table[row sep=crcr]{%
64	0.172183821102013\\
64	0.114127373472464\\
nan	nan\\
};
\addplot [color=mycolor3,solid,forget plot]
  table[row sep=crcr]{%
100	nan\\
100	nan\\
nan	nan\\
};
\addplot [color=mycolor1,solid,line width=1.0pt,mark=*,mark options={solid,fill=mycolor1}]
  table[row sep=crcr]{%
4	0.343901637678329\\
8	0.252648968468275\\
16	0.153671466464563\\
32	0.087364650404349\\
64	0.0751441681669349\\
100	0.0604301023475071\\
};
\addlegendentry{Our method};

\addplot [color=mycolor2,solid,line width=1.0pt,mark=square*,mark options={solid,fill=mycolor2}]
  table[row sep=crcr]{%
4	0.50721799226367\\
8	0.506164094763802\\
16	0.500499716706042\\
32	0.447455552832308\\
64	0.289368086247984\\
100	0.119744125730974\\
};
\addlegendentry{FITC};

\addplot [color=mycolor3,solid,line width=1.0pt,mark=triangle*,mark options={solid,fill=mycolor3}]
  table[row sep=crcr]{%
4	0.502773073223379\\
8	0.29778784064398\\
16	0.172286157081449\\
32	0.122560631964565\\
64	0.138024997424892\\
100	nan\\
};
\addlegendentry{VFF};

\end{axis}
\end{tikzpicture}%
  \caption{The effect of increasing the number of inducing inputs/features in the star-shaped domain regression study. The curves show the mean absolute error ($\pm$ std/min/max) in predictive mean compared to the results of the FULL GP model with noise-free observations along the boundary.}
  \label{fig:curve}
\end{figure}

For a quantitative evaluation, we compared estimates for all ten data sets and a differing number of inducing inputs/features. Fig.~\ref{fig:curve} shows the effect of increasing the number of inducing inputs/features and reports the mean absolute error (MAE) for the predictive mean compared to the results of the FULL GP. The results show that including the boundary information directly has clear benefits. The same conclusion can be drawn from  analysis of the marginal log-likelihoods and the predictive marginal variance estimates. In terms of computation time, our model is on par with VFF, with additional computational saving in evaluation of the marginal likelihood. Furthermore, the noise-free boundary measurements bring additional computational burden to the general-purpose schemes, while our model considers the boundary directly.

\subsection{Banana Classification Dataset}
\label{sec:banana}
We consider the banana classification dataset (see \cite{Hensman+Matthews+Ghahramani:2015}) with a hard decision boundary. We perform variational classification using a Gaussian approximation to the posterior $q(\vu)$ and optimise the ELBO with respect to the mean and variance of the approximation. We expect that increasing the number of harmonic features leads to an improved approximation; the variational framework guarantees that more inducing variables is monotonically better \cite{Titsias:2009}. However, this does not necessarily hold due to the restriction on $q(\vu)$.

\begin{figure*}[!t]
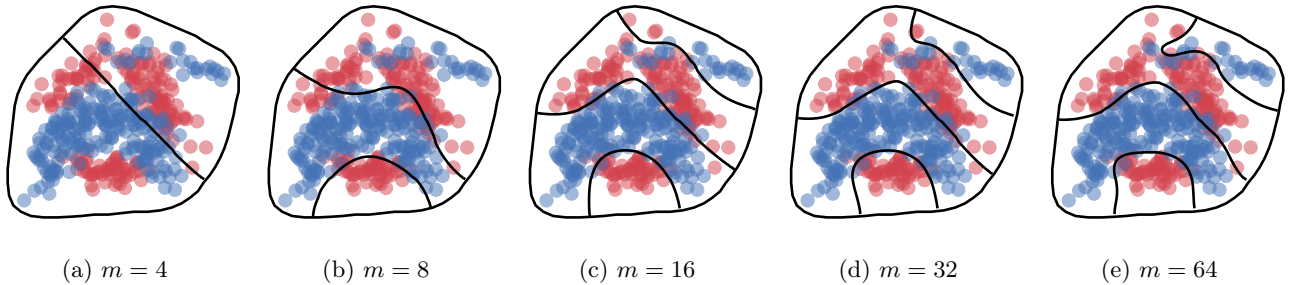

  \centering\footnotesize
  \pgfplotsset{yticklabel style={rotate=90}, ylabel style={yshift=-15pt},clip=false,scale only axis,axis on top,clip marker paths,legend style={row sep=0pt},legend columns=-1,xlabel near ticks,legend style={fill=white},axis line style={draw=none},tick style={draw=none}}
  \setlength{\figurewidth}{.19\textwidth}
  \setlength{\figureheight}{\figurewidth}
  \begin{subfigure}[b]{.19\textwidth}
    \input{./fig/banana-4.tex}
    \caption{$m = 4$}
  \end{subfigure}
  \hspace*{\fill}
  \begin{subfigure}[b]{.19\textwidth}
    \input{./fig/banana-8.tex}
    \caption{$m = 8$}
  \end{subfigure}
  \hspace*{\fill}
  \begin{subfigure}[b]{.19\textwidth}
    \input{./fig/banana-16.tex}
    \caption{$m = 16$}
  \end{subfigure}
  \hspace*{\fill}
  \begin{subfigure}[b]{.19\textwidth}
    \input{./fig/banana-32.tex}
    \caption{$m = 32$}
  \end{subfigure}
  \hspace*{\fill}
  \begin{subfigure}[b]{.19\textwidth}
    \input{./fig/banana-64.tex}
    \caption{$m = 64$}
  \end{subfigure}
  \caption{The effect of increasing the number of inducing features for the {\em banana} classification dataset with a hard decision boundary. In each pane, the coloured points represent training data and the decision boundaries are black lines. The outermost line is the pre-defined hard decision boundary.}
  \label{fig:banana}
\end{figure*}

Fig.~\ref{fig:banana} shows the two classes in the banana data set with red and blue markers. The pre-defined {\em hard decision boundary} is the solid black boundary enclosing all the data. It indicates the boundary outside of which the data does not play a role and at which there is absolute uncertainty of the correct class. We use a Bernoulli likelihood and a Mat\'ern ($\nu=\nicefrac{5}{2}$) kernel for the GP prior. We train the hyperparameters of the model jointly with the variational approximation. Fig.~\ref{fig:banana} shows the classification model outcomes with different numbers of inducing harmonic features. The black lines are the decision boundaries, which clearly improve with the increasing number of features.

\subsection{Tick Bite Density Estimation}
\label{sec:tick}
\begin{figure}[!t]
  \centering\footnotesize
  \pgfplotsset{yticklabel style={rotate=90},ylabel style={yshift=-15pt},clip=true,scale only axis,axis on top,clip marker paths,legend style={row sep=0pt},legend columns=-1,xlabel near ticks,legend style={fill=white},axis line style={draw=none},tick style={draw=none}}
  \setlength{\figurewidth}{.43\textwidth}
  \setlength{\figureheight}{1\figurewidth}
  \input{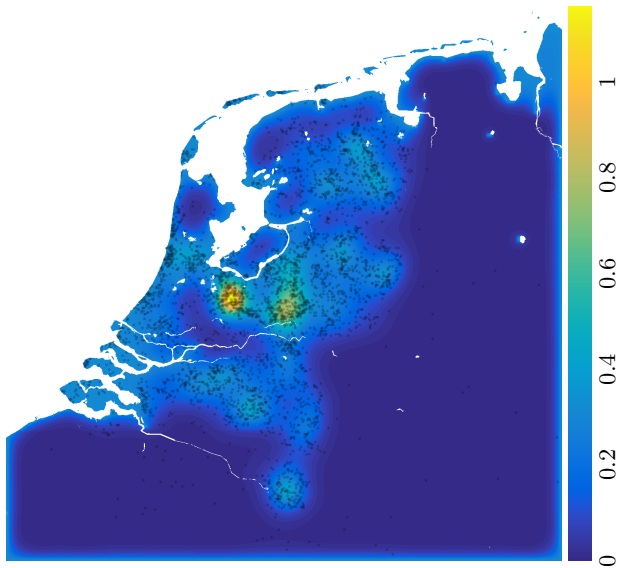} 
  \caption{Predicted density of tick bites per square kilometre around the Netherlands. The map is 400~km wide and the sea, rivers, and lakes (in white) are constraining the domain. The data (reported tick bites) are shown by the black dots, and they are modelled as a Log-Gaussian Cox process (Poisson likelihood).}
  \label{fig:tick}
\end{figure}

Ticks are small arachnids, typically 3--5~mm long. They are external parasites that live by feeding on blood typically of mammals and birds. Because of this, they may carry diseases that affect humans and other animals. To monitor the spread of ticks and of the diseases they might carry, many countries ask people to report tick bites. One such platforms is \url{https://tekenradar.nl} which is an initiative of the National Institute for Public Health and the Environment and Wageningen University, and collects data about tick bites in the Netherlands. The data is accessible online, and we used data of tick bites collected by this platform during the first 9~months of 2018. The 4,446 data points are scattered over the country as shown in Fig.~\ref{fig:tick}.

We use this data to model the tick density and exploit physical prior knowledge that ticks only live on land. The boundaries of our domain reflect this knowledge by assuming that the tick density is zero at sea, in the large lake in the middle of the country and in various rivers and lakes in the country. The area outside the domain in Fig.~\ref{fig:tick} is shown in white. We model the tick density using the number of ticks in a grid of roughly $n=15{,}000$ points and a Poisson likelihood in a Log-Gaussian Cox process model \cite{Moller+Syversveen+Waagepetersen:1998}. We used a Mat\'ern ($\nu=\nicefrac{3}{2}$) covariance function, and we train the hyperparameters of the model jointly with the variational approximation. The results are shown in Fig.~\ref{fig:tick}, where the colour indicates the intensity tick bites per square kilometre. As can be seen, this density explains the data well. For the eigendecomposition we used $m=256$ and a grid size of $200{\times}200$.

\section{DISCUSSION AND CONCLUSION}
\label{sec:discussion}
We have considered the problem of including physical knowledge about spatial constraints in GP inference. A numerical computation of the harmonic features of the GP prior allows us to specify boundary conditions on arbitrary shapes while at the same time attaining a low-rank representation that is used for speeding up inference. By approximating the posterior using variational inference, it is possible to use our method for non-Gaussian likelihoods. 

We have illustrated the efficacy of our method using both simulated and experimental data, with both Gaussian and non-Gaussian likelihoods. For the proposed method, there is a fixed setup cost for each new kind of domain, after which the method is as fast as the Hilbert GP  (and thus slightly faster than VFF). In Sec.~\ref{sec:benchmarking} ($m=100$) the setup cost was $4.9 {\pm} 0.1$~s, and for the banana example (Sec.~\ref{sec:banana} for $m=64$) the setup cost was $10.3 {\pm} 0.4$~s. After the preparation cost, the computation time for the actual GP inference is fast. Evaluating the marginal likelihood (or doing GP prediction) in Sec.~\ref{sec:benchmarking} takes 0.23~s when we have increased the number of observations to $n=10{,}000$ for the numbers to make more practical sense. If the variational approximation is used, most time is spent in the optimiser---as seen in the experiment in Sec.~\ref{sec:tick}, where the setup cost was in the range of some hundred seconds, but the optimiser ran for 25~min. This would, of course, be the same for all methods using the variational approximation.

The examples that we have presented only consider two-dimensional input domains. Higher-dimensional domains can, however, be considered in similar fashion as in \cite{Hensman+Durrande+Solin:2018}. Although the number of required basis functions grows exponentially in the input dimensionality $d$, Kronecker products and sums across the input dimensions have previously been used to address this problem. Throughout the paper we have focussed on using Dirichlet boundary conditions. The method can, however, be extended to be used with other boundary conditions, such as Neumann conditions. 

The main merit of our approach is that due to its simple and straightforward formulation, it can easily be extended or used in larger-scale systems. For example, the approach can  be extended to spatio-temporal analysis or be used in frameworks for simultaneous localisation and mapping (SLAM). The codes are available at \url{https://github.com/AaltoML/boundary-gp}.

\phantomsection%
\addcontentsline{toc}{section}{References}
\begingroup
\bibliography{bibliography}
\endgroup

\end{document}